\documentclass[conference,letterpaper]{IEEEtran}
\IEEEoverridecommandlockouts
\usepackage[T1]{fontenc}
\usepackage{cite}
\usepackage{amsmath,amssymb}
\usepackage{booktabs,array}
\usepackage{microtype}
\usepackage{xurl}
\usepackage[hidelinks]{hyperref}
\hypersetup{pdftitle={Benefit-Targeted Routing and Bounded Revision for LLM-Based Evidence Screening in Manufacturing Supplier Discovery},
 pdfauthor={Yijiashun Qi; Hanzhe Guo; Yuxuan Li}}
\newcommand{\yes}{\textit{yes}}
\newcommand{\no}{\textit{no}}
\newcommand{\unclear}{\textit{unclear}}
\newcommand{\doi}[1]{\href{https://doi.org/#1}{doi: \nolinkurl{#1}}}
\begin{document}
\title{Benefit-Targeted Routing and Bounded Revision for LLM-Based Evidence Screening in Manufacturing Supplier Discovery}
\author{
\IEEEauthorblockN{Yijiashun Qi\textsuperscript{*}}
\IEEEauthorblockA{University of Michigan\\
Ann Arbor, MI 48109, USA\\
elijahqi@umich.edu}
\and
\IEEEauthorblockN{Hanzhe Guo}
\IEEEauthorblockA{University of Michigan\\
Ann Arbor, MI 48109, USA\\
hanzheg@umich.edu}
\and
\IEEEauthorblockN{Yuxuan Li}
\IEEEauthorblockA{University of Pennsylvania\\
Philadelphia, PA 19104, USA\\
yuxuanli@alumni.upenn.edu}
\thanks{\textsuperscript{*}Corresponding author: Yijiashun Qi (elijahqi@umich.edu).}
}
\maketitle
\begin{abstract}
An evidence-review system must allocate additional model judgments and control their use. We study these software decisions on 6,164 frozen manufacturing source-task pairs. Benefit-targeted routing learns recovered agreement minus newly introduced disagreement. On the original 300-case test, 60 consultations yield 202/231 agreements with two human references versus 189/218 for a logistic-risk router. An additional 300 cases were independently annotated twice using equal frozen evidence. Without retraining the historical policies, the corresponding counts are 164/190 versus 163/193; the differences do not resolve an advantage. Strict scalar-Ridge controls share preprocessing, regularization and fitting, and remain inconclusive. Quotation and requested-state replacement gates retain rejected cases and budget costs; their joint gate rejects one original and two additional consultations without changing primary outcomes. Evidence ablations, fusion and bounded updates supply further controls. A completed 200-pair identity audit yields 97.45 percent weighted merge precision after eight author adjudications. This audits a provisional map, while prospective company-disjoint performance remains untested.
\end{abstract}
\noindent{\footnotesize\textit{Preprint -- October 18, 2026.}}\par

\section{Introduction}
A difficult supplier-screening record is not necessarily a useful target for another model. The additional reviewer may repeat the first judgment or replace a reference-consistent answer with a worse one. A review pipeline consequently needs to distinguish three decisions: acquiring another judgment, combining judgments already available, and applying a revision to a structured record. Optimizing these decisions against different targets can produce conflicting outcomes.

We study this computational problem on fixed web evidence for manufacturing supplier discovery. Our main intervention routes additional review by predicted \emph{paired benefit}: the expected improvement from a particular replacement, including its potential harm and unavailable-response fallback. This differs from prioritizing the probability that the delivered answer disagrees with a human reference. We compare the two strategies with identical test-time information available before consultation, and separately test whether richer evidence representations or post-consultation fusion improve results.

The contributions are a failure-inclusive consultation and replacement implementation; objective, quality-gate and additional-reference controls that test its supported advantage; and a bounded field-update contract separating processing guarantees from reference quality. Negative controls are retained. These connect acquisition, aggregation and revision to measurable software behavior, without claiming a new regression or voting algorithm.

Our questions are: \textbf{RQ1}, when does benefit-targeted consultation outperform risk-based review? \textbf{RQ2}, what extra value comes from evidence text and a second model's outputs? \textbf{RQ3}, how does enforcing revision scope affect processing and reference concordance? The study uses standard statistical learners to test these design choices, with all new configurations fixed before their current fits.

\subsection{Related Work}
RouteLLM learns routing from preferences,\cite{ong2025} LLM-Blender ranks and fuses responses,\cite{jiang2023} and learning-to-defer incorporates downstream capabilities.\cite{mozannar2020} We test the net effect of replacing a delivered categorical judgment, retaining failed consultations in the budget. Adaptive pruning in mixture-of-agents serving,\cite{wang2026serving} layer-skipping for vision-language-action inference,\cite{yang2026vla} and embodied-system resilience\cite{xie2026resilience} provide broader allocation and failure-aware context. SPADE studies pipeline assertions,\cite{spade2024} and AFaCTA uses reasoning-path consistency for annotation confidence.\cite{ni2024} Our controls distinguish pre-consultation ranking, returned-output validation and permitted update scope.

Manufacturing capability prediction and supplier retrieval already use graphs,\cite{li2024,li2025} including SME funding-progression modeling\cite{qisme2026} and coverage-aware web acquisition.\cite{qi2026} These motivate the application but answer different questions from fixed-evidence review. Search-source differences,\cite{yagci2022} judge biases,\cite{zheng2023} citation quality,\cite{gao2023} and sufficient-context analysis\cite{joren2025} motivate explicit evidence and evaluation boundaries. Related work on correlated memory promotion\cite{qimemory2026} and negation repair\cite{qihints2026} similarly cautions that another review or correction can introduce error.

\section{Data and Evaluation Design}
\subsection{Frozen Evidence and Model Archives}
On July 30, 2026, 40 capabilities, eight states and three intents formed 960 queries. Top-ten Brave/You requests returned 9,590/9,600 rows, with one failed Brave query. Independently fetched publisher pages yielded 6,164 source-capability-state pairs; provider payloads were not retained. A pair is not a unique firm, and query wording does not verify SME size.

Model inputs contain task, title, description and keyword-selected source windows under a 9,000-character budget, without browsing. Labels \yes/\no/\unclear\ express support for business identity, direct production, requested capability, in-state production and commercial offering. Component enums, exclusion reasons and quotations support review; offices and service areas do not establish production.

GPT-5.6-Sol supplies framed passes A/B, C adjudication on ten-field disagreements, and subsequent quotation re-review. GLM-5.3 independently receives the same evidence without GPT labels, at \texttt{output\_config.effort=max}. Its 309 batches require 317 wire requests under a documented retry amendment: 5,893 valid rows, one invalid, 50 omitted, 220 failed and none unattempted. The earlier partial run is excluded. Valid positive flags comprise 73 nonliteral quotations and five state mismatches. GPT/GLM concordance is 4,569/5,893 (77.53 percent), $\kappa=.619$, with 1,074 joint positives. Unequal call histories confound model/workflow effects.

\subsection{Human References and Isolation}
Original H1/H2 contain 1,200 first and 300 independent second annotations, with participation confirmed by the first author. Reviewers initially saw neither model labels nor each other's answers but could use supplemental sources. Primary agreement is 244/300 (81.33 percent), $\kappa=.706$; all 56 differences involve \unclear. Component agreement spans 86.00--91.67 percent.\cite{artstein2008} First-author self-review later addresses 94 disputes, 44 using supplemental evidence. We retain original references rather than treating this composite as independent equal-evidence arbitration.

Original learned experiments use 300 test cases across 256 recorded domains. Removing 273 of the other 900 cases sharing a test domain/hash leaves 507 training and 120 validation cases, with 627 for final fitting. Recorded domain/hash intersections are zero; aliases may remain. H1 provides development targets; H2 supplies sensitivity evaluation. Human notes, supplemental sources and composite labels never enter features.

Additional validation samples 300 cases outside the original 1,200, one per screened evidence component, across 300 domains. Components link domain, nonempty hash, eligible normalized title/excerpt, or five-word-shingle similarity: Jaccard $\geq.80$ with 50 shingles, or containment $\geq.90$ with 150. Components touching old inputs are excluded. All 22 empty excerpts remain. Two independently blinded annotators use frozen evidence only, with participation confirmed by the first author. J1/J2 agree on 240/300 labels (80.00 percent), $\kappa=.655$; 60 primary and 92 primary/component disputes remain unadjudicated. Company-level isolation is not verified.

A separate identity audit double-labels 200 distinct-source pairs from 1,072 source units in the provisional positive map: 100 proposed merges and 100 non-merges. Presampled merge strata cover all 28 cross-domain and 22 name-change pairs, plus 50 of 459 same-name pairs; non-merge strata sample 50 alias-risk and 50 other pairs. Both humans start blank with proposals hidden and randomized presentation. The first author resolves eight relationship disputes, declaring method participation and model exposure. Original independent answers remain unchanged; this is author adjudication, not a third independent reference.

The original test was examined earlier. Earlier runners fingerprint predictions before decoding their test references, but remain retrospective. Historical policy weights were frozen October 9; supplementary inference does not refit them. New reference answers had been inspected at intake before the follow-up quality gates and strict-control protocol were fixed; these additions are exploratory. We use 2,000 paired recorded-domain bootstrap replicates on fixed predictions: original routing intervals are 97.5 percent, evidence/fusion families 98.75 percent, and the follow-up's six primary contrasts per dataset 99.17 percent. These adjustments do not cover all research iterations, training uncertainty or unresolved aliases.

\section{Review Decisions and Implementations}
\subsection{Benefit-Targeted Consultation}
Let $g_i$ be the delivered GPT label, $z_i$ a schema-valid GLM label, and $h_i$ the development human reference. Define $\tilde z_i=z_i$ when available and $\tilde z_i=g_i$ otherwise. The supervised paired outcome is
\begingroup
\begin{equation}
 d_i=\mathbf{1}[\tilde z_i=h_i]-\mathbf{1}[g_i=h_i]\in\{-1,0,1\}.
 \label{eq:gain}
\end{equation}
\endgroup
Thus an unavailable result contributes zero, and a harmful replacement contributes $-1$. Under additive equal-record utility, maximizing expected improvement for exactly $B$ consultations ranks records by $\mathbb{E}[d_i\mid x_i]$. Initial disagreement risk, $\Pr(g_i\ne h_i\mid x_i)$, omits whether the specific alternative can help or harm. These are different prediction targets.

Pre-call inputs comprise training-fitted TF-IDF/SVD64, 40 capability/eight state indicators, and 27 GPT channels: A/B/final labels, A/B categorical confidence, ten disagreement bits, their count and C-routing. GLM outputs/availability/flags are forbidden. Training-only constant removal and scaling precede Ridge; $\alpha\in\{.1,1,10\}$ minimizes validation MSE with larger-$\alpha$ ties. Selected $\alpha=.1$ refits on 627 cases; scores and hash tie order are frozen.

The prior logistic-risk router uses the same 139 pre-call features, with validation log-loss selection. Estimator and objective both differ; paired-benefit training additionally uses archived GLM outcomes. Baselines retain GPT, always consult GLM with fallback, use random expectation, prioritize \unclear, or prioritize A/B primary disagreement.

Exploratory matched-family controls add risk Ridge ($\alpha=10$) and three-class gain logistic ($C=.1$), scored by $P(d=1)-P(d=-1)$. They retain identical preprocessing and prior rankings. Their four reference contrasts use 98.75 percent intervals; selected regularization and logistic output dimensions differ.

Strict controls instead fit two scalar Ridge models to the same development matrix, with common $\alpha=1$, SVD solver and fitting procedure, changing only risk/gain targets. No new parameter search or J1/J2 training occurs. Additional predictions are fingerprinted before this runner computes metrics; prior reference exposure remains disclosed.

Exactly $B$ records are selected before archived GLM replacement. Unavailable results consume budget and retain GPT. Primary $B=60$, with secondary 30/90, counts planned record evaluations, not batches, money or latency. Changed acquisition context could invalidate the fixed-response replay assumption.

Post-call replacement compares schema-only, nonliteral-positive-quotation, positive requested-state mismatch and joint gates, using verified archival flags. Rejected responses retain GPT and consume budget. Occurrence supports traceability, not entailment; matching returned state does not verify a production facility.

\subsection{Evidence Representation and Post-Consultation Fusion}
Six logistic arms compare task indicators; title/description plus task; excerpt plus task; full text plus task; full text without task; and sparse full-text TF-IDF plus task. TF-IDF uses at most 8,000 unigram/bigram features, minimum frequency two, sublinear counts and L2 normalization; compressed arms use SVD64. Empty text remains zero. Validation log loss selects $C\in\{.1,1,10\}$: .1 except text without task, which selects 1.

Post-call fusion adds 30 GLM label/component/availability channels to the 139-channel GPT baseline; GLM confidence is absent and not invented. Separate logistic models predict H1 class or GPT/H1 disagreement, selecting the same $C$ grid by validation log loss. Unavailable GLM exactly inherits baseline probabilities/scores, including validation. All 300 cases remain primary; 282 valid-return cases are secondary. Quality flags do not select favorable cases.

\subsection{Bounded Updates to Structured Records}
An adjudicator can alter fields unrelated to escalation. For ten key fields $F$, let $D_i=\{f:a_{if}\ne b_{if}\}$. Whole-record replacement selects C whenever $D_i$ is nonempty. Fieldwise selection is
\begingroup
\begin{equation}
 m_{if}=\begin{cases}a_{if},&a_{if}=b_{if},\\c_{if},&a_{if}\ne b_{if},\end{cases}
 \label{eq:merge}
\end{equation}
\endgroup
which preserves A/B-agreed key fields in $O(n|F|)$ selections. This equals per-field A/B/C plurality with all-different ties resolved by C; its role is enforcing the allowed update scope. Non-key rationale/quotation fields remain outside that guarantee. A separate check aborts incompatible positive records. Later quotation re-review may explicitly revise a decision; literal normalized occurrence checks traceability, not entailment. Paired replay uses identical saved responses, isolating merge behavior from new inference.

\section{Experimental Results}
\subsection{RQ1: A Useful Second Review Is Action-Dependent}
At budget 60, the benefit policy attempts 60 consultations, receives 56 valid judgments, and retains GPT for four unavailable cases. Twenty-one labels switch. Against H1, 17 changes gain agreement and three lose it, raising concordance from 188/300 to 202/300. H2 gains 18 and loses three, increasing 216/300 to 231/300. The old risk router gains nine/ten but loses eight under each reference, reaching only 189/218 (Table~\ref{tab:routing}).

\begin{table}[t]

\caption{Fixed-response routing on all 300 cases. G/L: gained/lost reference agreements. Random rows are exact count expectations. $\dagger$ marks the later exploratory estimator-family controls.}
\label{tab:routing}\centering
\begin{tabular}{@{}lrrrr@{}}
\toprule
Policy & Attempts & Valid & H1 G/L & H2 G/L\\
\midrule
GPT only & 0 & 0 & 0/0 & 0/0\\
Random expectation & 60 & 56.4 & 6.6/3.2 & 6.0/4.2\\
A/B disagreement & 60 & 54 & 11/6 & 12/5\\
Unclear first & 60 & 57 & 13/6 & 15/5\\
Risk, logistic & 60 & 57 & 9/8 & 10/8\\
Benefit, Ridge & 60 & 56 & 17/3 & 18/3\\
Risk, Ridge$\dagger$ & 60 & 57 & 10/8 & 13/6\\
Benefit, logistic$\dagger$ & 60 & 56 & 17/7 & 18/7\\
Always GLM, fallback & 300 & 282 & 33/16 & 30/21\\
\bottomrule
\end{tabular}
\end{table}

The original benefit-minus-risk difference is $+13$ per reference, with 97.5 percent intervals $[1,22]$/$[1,20]$, meeting that original comparator-specific criterion. Macro-F1 is .6672/.7294 versus .6236/.6773. Expanded follow-up intervals at 99.17 percent are [$-1,24$]/[$-1,22$], crossing zero. Exploratory 95 percent intervals versus \unclear-first also cross zero: H1 $+7$ [$-4,13$], H2 $+5$ [$-5,10$]. Always GLM yields 205/225 agreements. At budgets 30/90, benefit-minus-risk differences are 8/5 (H1), 7/5 (H2), with intervals touching or crossing zero. The result is not uniform across comparisons or budgets, and does not measure acquisition-cost savings.

The within-family follow-up preserves the direction but does not resolve the advantage. At 60, benefit minus risk within Ridge is $+12$ (H1, 98.75 percent interval [$-1.51,19$]) and $+8$ (H2, [$-3,17$]); within logistic it is $+9$ for both, with intervals [$-2,21$] and [$-2.51,20$]. All include zero. Thus the original cross-family result supports that specific strategy comparison, while a target-specific or estimator-independent superiority claim remains unestablished.

Additional-reference validation does not reproduce a resolved advantage (Table~\ref{tab:additional}). Frozen benefit minus frozen risk at 60 is $+1$ (J1, 99.17 percent interval [$-10,12.67$]) and $-3$ (J2, [$-14,9$]). Strict common-Ridge differences are $-2$ [$-13,10$] and $-4$ [$-15,8.67$]; on the original pool they are $+10/+8$, with intervals [$-2,19$]/[$-4,18$]. All cross zero. The original $+13/+13$ comparison is therefore dataset- and comparator-specific.

\begin{table}[t]

\caption{Additional 300 cases: concordant primary labels at $B=60$. J1/J2 are the separate new human references. Common Ridge uses shared $\alpha=1$. No test-dependent refitting.}
\label{tab:additional}\centering
\begin{tabular}{@{}lrr@{}}
\toprule
Policy & J1 & J2\\
\midrule
Unchanged GPT ($B=0$) & 161 & 190\\
Frozen risk, logistic & 163 & 193\\
Frozen benefit, Ridge & 164 & 190\\
Frozen risk, Ridge & 165 & 193\\
Frozen benefit, logistic & 160 & 188\\
Common-Ridge risk & 166 & 194\\
Common-Ridge benefit & 164 & 190\\
\unclear-first & 150 & 179\\
A/B disagreement-first & 155 & 183\\
Always GLM ($B=300$) & 182 & 202\\
\bottomrule
\end{tabular}
\end{table}

GLM returns are valid for 292 additional cases, with seven failures and one omission; five valid positives have quotation flags and one a state flag. At 60, benefit routing receives 59 valid returns with one unavailable fallback. Quote/state gates reject one each, jointly two; the original pool jointly rejects one. These rejected primary labels already equal delivered GPT, so switches, gains, harms, concordance and macro-F1 remain unchanged. Under always-consult diagnostics, the joint gate rejects three original and six additional records; original concordance falls from 205/225 to 204/224, while additional concordance stays 182/202. Gating has not demonstrated a quality gain; it can also remove reference-consistent judgments.

\subsection{RQ2: More Information Does Not Guarantee Better Use}
Complete text plus task beats task alone by macro-F1 .0396 (H1, 98.75 percent interval [.0048,.0785]) and .0417 (H2, [.0071,.0810]). It does not beat title/description plus task: differences are $-.0079$ [$-.0448,.0284$] and $-.0100$ [$-.0472,.0259$]. Median exposed characters increase from 215 for headers to 5,119 for full text. Table~\ref{tab:evidence} includes the sparse control. Most arms fail to predict \unclear; this limits conclusions about the evidence itself. The text-without-task arm has the highest point estimates but a different selected regularization, so its exploratory advantage does not isolate a harmful task effect.

\begin{table}[t]

\caption{Evidence ablations: three-class macro-F1 against original references; positives among the highest 60 positive scores. T denotes capability/state indicators.}
\label{tab:evidence}\centering
\begin{tabular}{@{}lrrr@{}}
\toprule
Representation & H1 F1 & H2 F1 & H1/H2 @60\\
\midrule
T only & .2971 & .3410 & 29/29\\
Title + description + T & .3445 & .3927 & 29/30\\
Excerpt + T & .3230 & .3691 & 31/32\\
Full text + T & .3366 & .3827 & 30/31\\
Full text, without T & .3922 & .4600 & 33/34\\
Full sparse TF-IDF + T & .3146 & .3582 & 32/33\\
\bottomrule
\end{tabular}
\end{table}

Post-consultation fusion raises macro-F1 point estimates from .6037/.6488 to .6206/.6888, but the four primary classification/ranking intervals include zero (Table~\ref{tab:fusion}). Classification differences are .0168 [$-.0339,.0719$] and .0400 [$-.0318,.1082$]. Risk-ranking yield at 60 falls from 36/38 to 34/35, differences $-2$ [$-8,7$] and $-3$ [$-10,5$]. Thus the returned GLM information has not established incremental learned-fusion value here. Raw GLM with GPT fallback scores .6821/.7151, a useful diagnostic whose historical inference workload differs from GPT's.

\begin{table}[t]

\caption{Post-consultation comparison on all 300 cases. F1: primary classification; D@60: human-reference disagreements with unchanged final GPT found by separately trained risk models. Unavailable fusion inputs retain matched GPT baseline outputs.}
\label{tab:fusion}\centering
\begin{tabular}{@{}lrrrr@{}}
\toprule
 & \multicolumn{2}{c}{Macro-F1} & \multicolumn{2}{c}{D@60}\\
Method & H1 & H2 & H1 & H2\\
\midrule
Learned GPT strong & .6037 & .6488 & 36 & 38\\
Learned GPT + GLM & .6206 & .6888 & 34 & 35\\
\bottomrule
\end{tabular}
\end{table}

\subsection{RQ3: Processing Guarantees and Quality Differ}
C reviews 3,292 records, including 2,145 with an agreed primary label but another disputed key field. Whole-record replacement changes 321/2,145 agreed primary labels (14.97 percent), all \no\ to \unclear; fieldwise selection changes none. Any agreed-key overwrite occurs in 339/6,164 records under whole-record selection and zero under the bounded contract. Independently reconstructing the final pipeline matches all 98,624 stored decision-field values. The subsequent quotation stage reviews 71 provisional positives and changes 31 to \unclear, yielding final counts of 1,148 \yes, 3,799 \no, and 1,217 \unclear.

On the 300 human-reviewed cases, only eight labels differ between merge policies. Fieldwise selection gains four and loses four H1 agreements (191 under each policy), and gains five and loses three H2 agreements (217 to 219). Macro-F1 differences are $-.0065$/$-.0029$, with exploratory 95 percent intervals including zero. Preserving agreed fields enforces update scope but can preserve shared mistakes. The separate quotation stage changes three test positives to \unclear\ despite both human references being \yes; their broader evidence access prevents interpreting this as proven erroneous correction.

Prior diagnostics also remain inconclusive: trace features find 35/36 disagreements at 60 versus 36/38; component-necessity regularization\cite{xu2018} raises macro-F1 .0211/.0132 with intervals including zero. Hard gating removes incompatible positives while reducing recall. GraphSAGE\cite{hamilton2017} finds 28/29 positives versus 31/32 for text. Full settings and negative results remain in the aggregate report.

\section{Discussion and Conclusion}
The implemented pipeline separates allocating consultation, validating a replacement and bounding update scope. Its original benefit-routing advantage does not resolve under strict controls or additional equal-evidence references. Mechanical quotation/state gates also do not guarantee better concordance. These controls narrow the contribution to a reproducible failure-inclusive implementation and its empirical limits, while the bounded-update contract independently prevents out-of-scope changes.

The completed identity audit has operating/group agreement of 196/200 and 195/200 ($\kappa=.962/.950$). Population-weighted merge precision is 97.25/97.05 percent under the original references and 97.45 percent after author adjudication, with unknowns counted conservatively. The composite confirms 22/28 cross-domain, 15/22 name-change and 50/50 same-name merges; 11 proposed merges are rejected and two remain unclear. Non-merge correctness is 96.00 percent; six operating and 100 corporate relations remain unclear overall. The prespecified bootstrap degenerates on census/all-success strata. A disclosed post-intake finite-population diagnostic gives a conditional 95 percent interval of [91.55,97.45] percent for composite merge precision, not a 95 percent lower-bound guarantee.

Limitations include prior reference exposure, multiple experimental families, one application, unequal inference histories, and different human evidence access. The identity audit covers the provisional positive map, not full-corpus company resolution; 300 recorded domains are not 300 verified companies. Company-disjoint performance and a reserved test after all revised policies are locked remain needed. We do not claim these requirements are fulfilled, universal routing superiority, factual supplier accuracy or measured acquisition-cost savings.

\section*{Reproducibility and AI Assistance}
Model-audit code and aggregates are at \url{https://github.com/elijahqi/us-sme-signal-graph} (release anchor \texttt{c036829}). New protocols, frozen predictions, tests and row-free summaries are locally archived; raw annotations and third-party text remain private. GPT-5.6-Sol and GLM-5.3 supplied experimental judgments. AI tools assisted implementation and writing under author direction; the authors remain responsible for verification. Additional human participation, blinding and practice were confirmed by the first author; file timestamps do not establish reading time. Original exports and metadata discrepancies are preserved separately.

\end{document}